# Spotting prosodic boundaries in continuous speech in French

*Vincent Pagel* [(1)], *Noëlle Carbonell* [(1)], *Yves Laprie* [(1)], *Jacqueline Vaissière* [(2)]
(1) CRIN, BP 239, 54506 Vandoeuvre-Lès-Nancy, France
(2) ILPGA, 19 Rue des Bernardins, 75006 Paris, France

**ABSTRACT**

A radio speech corpus of 9mn has been prosodically marked by a phonetician expert, and non expert listeners. This corpus is large enough to train and test an automatic boundary spotting system, namely a time delay neural network fed with F0 values, vowels and pseudo-syllable durations. Results validate both prosodic marking and automatic spotting of prosodic events.

**CONTEXT AND MOTIVATION**

It is known for a number of languages that speech contains prosodic cues acting as boundary markers of different strength along the continuum. Boundary marking is particularly obvious in French, which has no distinctive lexical stress. Fundamental frequency (F0) movements are generally bounded by left and right word boundaries and phonemic lengthening marks the end of the sense groups. Besides, prominence is usually achieved through accents (F0 rises mostly) on monosyllables and on the first syllables of polysyllables. However, it is not clear whether and how prosodic cues may be used for segmenting continuous speech automatically.

Previous research using heuristic rules in expert systems [1][2], has uncovered problems, due mainly to: the diversity of intrinsic phonemic durations (nasal vowels are longer); the effects of the rate of speech (fewer and less obvious boundaries in rapid speech); inter-speaker variations; and the weighting of F0- vs. duration cues.

Moreover, in situations that favor expressiveness, accents may be misinterpreted as right word boundaries. This explains why current research on the automatic segmentation of speech into prosodic units applies to read speech only, namely to the exclusion of spontaneous oral communication where the expressive function of prosody prevails against its linguistic one.

We are currently studying «controlled speech», e.g. radio news announcements and press reviews, with a view to extending the scope of continuous speech recognition applications. The prosodic processing of «controlled speech» should prove easier than the analysis of spontaneous speech, since newscasters aim at and achieve balanced trade-offs between expressive and communicative purposes.

**OBJECTIVES AND METHOD**

The paper investigates the two following issues:
• Which acoustic parameters should be selected in order to discriminate left from right word/group boundaries accurately?
• Is the prosodic coding scheme we use consistent enough?

To answer these questions, we tested a multi-layer perceptron on a «controlled speech» corpus, using different sets of prosodic marks for the training stage.
Nine minutes of a radio press review, spoken by a single speaker, were phonetically labeled by a phonetician and prosodically coded both by a group of listeners and by an expert on prosody (J. Vaissière).
Twenty French phonetics students listened twice to the press review. They were asked to jot down (on the fly), first,

the prosodic group end-boundaries they noticed (first audition), then the syllables they perceived as accented (second audition).

The expert coded F0 movements from a visual representation of the acoustic-phonetic data made up of: the phonetic segmentation marks and labels, the smoothed curve of F0, vocalic and intervocalic duration curves (all time-aligned and on the same sheet) computed from the phonetic segmentation. The wide-band spectrogram was also available but on a separate sheet.

**ACOUSTIC-PROSODIC CODING**

The expert described meaningful F0 movements and pauses, using a TOBI-like coding scheme we developed for French. Our coding symbols are presented in table 1; capital letters describe major F0 movements; indexes are used to indicate the position of the F0 movement inside the current word or prosodic group; symbols and indexes can be combined. For instance, the initial F0 rise in the current prosodic group is coded R- when it occurs on the first syllable of the current word; B+Rc indicates a crossing of the baseline followed by a continuation rise (at the end of the current prosodic group).

**Results of the coding**

The marks from 3 listeners were excluded from the analysis, since these subjects had difficulty in detecting accents.

The locations of the end-boundaries and accents perceived by the 17 remaining subjects were compared to both the expert's coding (F0 movement) and the computed vocalic and intervocalic durations, in order to:

-determine which prosodic phenomena characterize perceived word or group boundaries, and then identify efficient acoustic input data for the MLP;

- evaluate the size of the optimum context in terms of the number of sounds on the right and/or on the left of the marked syllable.

We shall not comment on the distribution and acoustic correlates of perceived word or group end-boundaries, because subjects' judgements agree with each other, and most syllables marked by them are followed by pauses which can be reliably detected by standard algorithms. Nevertheless, lengthening prevails over pauses and F0 specific contours as a decision factor.

*Table 1. Our prosodic coding symbols*

| **R** : initial rise on the first syllable of a word | **Ri, Li** : movement delayed on the i*th* syllable |
|---|---|
| **L** : prominent fall on the last syllable of a word | **R-, L-** : on the head or tail of the current group |

| **P** : peak | **P^** symmetric slopes | **P/** left slope deeper than the right one |
|---|---|---|
| | **Ph** particularly high F0 values | **P\\** right slope deeper than the left one |

| **B** crossing of the baseline |
|---|
| **Rc** continuation rise (last syllable of the prosodic group |
| **S** «sustained» (last syllable of prosodic groups) |
| **U** valley on a grammatical word (between two prosodic groups) |
| **V** sharp dip due to enhanced micromelody (separates two words) |

On the contrary, the distribution of perceived accents provides useful knowledge. 80 syllables were perceived as accented by more than 2 subjects (i.e. 1 perceived accent per 5,2 sec. time interval on the average). 10% of the perceived accents are syllables unmarked by the expert, half of them following a U mark; which suggests a rather limited influence of meaning on the perception of accents. The acoustic correlates for 5% of the marked syllables are atypical (for instance, F0 on the baseline), while the analysis of the other marks confirms previous studies: F0 (cf. Table 2) is the major cue for detecting prominence which may affect even grammatical words (16%); polysyllables are usually accented on the first syllable (85%), which is typical of news announcers' styles; lengthening is optional; when present (57%), it is moderate (compared to group end-boundaries) and affects consonants (64%) rather than vowels which may be shortened (10%).

*Table 2. F0 movements corresponding to perceived accents*

| expert mark | mono-syllable | first syllable | last syllable |
|---|---|---|---|
| **P** | 9%[a] | 1% | 5% |
| **R-** | 11%[b] | 11% | |
| **R** | 15% | 9% | |
| **Total** | 35%[c] | 26% | |

a. 0% on grammatical words
b. 9% on grammatical words
c. Grand total reckon for 61% of the marks because other perceived marks where not coded by the expert

Besides, listeners' judgements favor unexpected phenomena against regularities: F0 peaks on the last syllables of polysyllables are generally ignored, as well as peaks on the second or third syllable; the first peak in a sequence of peaks (cf. digit sequences) is marked, while the following ones are not perceived as accents, even if they are more prominent than the first one.

These results indicate that local prosodic events provide useful reliable linguistic information on word and group boundaries, on condition that the interpretation of local phenomena involve contextual information on the long-term evolution of prosodic parameters

**Prosodic segmentation using MLP**

In the following, we use MLPs, implemented with cross-validation to avoid over-training, and with the softmax transfer function so that we get *maximum a posteriori probabilities* (MAPs) as described in [3]. When the training subset is not balanced, MAPs are divided by a priori probabilities for each class we want to recognize, so that we get scaled likelihoods. We decided that the system answers if one likelihood is greater than the sum of all the other likelihoods, so it is possible that the system gives no answer for a given input.

For each test, we use the last 75% of the speech corpus to train the MLP, and we perform the test on the first 25%.

We tried several inputs combinations as well as their derivatives: F0 average and regression coefficient on a vocalic segment, segmental duration, and pseudo-syllable duration. This last parameter is the time elapsed between the end of a vowel and the end of the next one, because in French the CV-CV syllable scheme is encountered most of the time. Note that we are not exactly in a true speech recognition situation, as the phonetic labeling gives vowels positions, but we do not consider it as a handicap since there exist reliable vocalic nucleus detectors nowadays.

### Auditory marks

Two kinds of marks have been set by the listeners *(frontiers* and *accents)*, which are attached to the syllable nucleus.

The MLP fed with any of the previously described values (F0,duration...), no matter the size of the temporal window, is not capable of reproducing the *accent* marking with a good score. Thus we consider that listeners' accent marks are not consistent, at least from a local point of view.

But for the *frontier* marks, the MLP fed with the duration, on a 5 vowel context, achieves the task with 11% insertion and 43% omissions.

### Phonetician marks

At this stage, we use the auditory marks to select a significative subset of marks set by the expert. Considering the given number of mark types obtained, we found it necessary to gather them in generic classes to achieve a correct training of the MLP : R for initial rise (129 occurrences), P for peaks (128), B for baseline (105), C for continuation rise (50), Nil for no marking at all (1287).

*Table 3. Confusion matrix: horizontally, expected results, vertically, MLP results. (356 answers / 400)*

|     | Nil | B  | C | P  | R  |
|-----|-----|----|---|----|----|
| Nil | 227 | 6  | 0 | 4  | 3  |
| B   | 7   | 20 | 0 | 0  | 0  |
| C   | 0   | 5  | 7 | 1  | 1  |
| P   | 5   | 0  | 1 | 25 | 5  |
| R   | 0   | 0  | 0 | 5  | 34 |

After several tests, we kept vowel duration, F0 values, and pseudo-syllable duration on a 7 vocalic nucleus window to feed a MLP with 10 neurons in its hidden layer. The MLP has 5 outputs: one for each class mentioned above.

The MLP gives no answer for 44 configurations (concurrent answers). Surprisingly, no nasality tag is required to draw the MLP attention on the fact that nasal vowels are much longer than vocalic ones.

## RESULTS AND CONCLUSION

The main result is that this experience validates both the expert prosodic marking and the automatic spotting system. Furthermore, the confusion rate between P and R marks is rather low, which agrees with the results of [4]: lengthening is a more important correlate of F0 peak for P than for R. R marks recognized as P, are accented monosyllabics words.

The recognition rate for C is enhanced when we add F0 regression parameters, as involved vowels bear a long upward F0 move. However this adds a slight confusion in the identification of P marks.

Future work will aim at incorporating long term prosodic variations in the modelling of our prosodic marks.